\documentclass[conference]{IEEEtran}
\IEEEoverridecommandlockouts
\usepackage{cite}
\usepackage{amsmath,amssymb,amsfonts}
\usepackage{algorithmic}
\usepackage{ulem}
\usepackage{amsmath}
\usepackage{amsthm}
\usepackage{graphicx}
\usepackage{subfigure}
\usepackage{color}
\usepackage{graphicx}
\usepackage{textcomp}
\usepackage{multirow}
\usepackage{booktabs}
\usepackage{threeparttable}
\usepackage{xcolor}

\usepackage[marginal]{footmisc}

\def\BibTeX{{\rm B\kern-.05em{\sc i\kern-.025em b}\kern-.08em
		T\kern-.1667em\lower.7ex\hbox{E}\kern-.125emX}}
\begin{document}
	
	\title{Single Image Deraining via Rain-Steaks Aware Deep Convolutional Neural Network\\
		{\footnotesize \textsuperscript{}}
	}
	
	\author{\IEEEauthorblockN{Chaobing Zheng}
		\IEEEauthorblockA{School of Information \\ Science and Engineering, \\
			Wuhan University of \\ Science and Technology,\\
			Wuhan, China \\
			zhengchaobing@wust.edu.cn}
		\and
			\IEEEauthorblockN{Yuwen Li$^{*}$}
			\IEEEauthorblockA{School of Information \\ Science and Engineering, \\
			Wuhan University of \\ Science and Technology,\\
			Wuhan, China \\
			liyuwen1205@163.com}
		\and
		\IEEEauthorblockN{Shiqian Wu \dag}
		\IEEEauthorblockA{School of Information  \\ Science and Engineering, \\
			Wuhan University of \\ Science and Technology,\\
			Wuhan, China \\
			shiqian.wu@wust.edu.cn}
	}
	
	\maketitle

\footnote{${*}$ contributing equally, \dag corresponding author    }

\begin{abstract}
It is challenging to remove rain-steaks from a single rainy image because the rain steaks are spatially varying in the rainy image. This problem is studied in this paper by combining conventional image processing techniques and deep learning based techniques. An improved weighted guided image filter (iWGIF) is proposed to extract high frequency information from a rainy image. The high frequency information mainly includes rain steaks and noise, and it can guide the rain steaks aware deep convolutional neural network (RSADCNN) to pay more attention to rain steaks. The efficiency and explain-ability of RSADNN are improved. Experiments show that the proposed algorithm significantly outperforms state-of-the-art methods on both synthetic and real-world images in terms of both qualitative and quantitative measures. It is useful for autonomous navigation in raining conditions.
\end{abstract}

\section{Introduction}
Despite  broad applications of outdoor vision systems, existing systems do not account for different weather conditions \cite{1Nara2000}. Based on differences in their physical properties, weather conditions are usually classified as static (fog, mist and haze) or dynamic (rain, snow and hail) \cite{1greg2004}. Volumetric scattering models such as attenuation and airlight can be used to adequately describe the effects of steady weather \cite{1Nara2000,2li2021}. The analysis of dynamic weather conditions requires the development of stochastic models that capture the spatial and temporal effects \cite{1greg2004}. In order to develop vision systems that perform under all weather conditions, it is essential to model the visual effects of the various weather conditions and develop algorithms to remove them \cite{zheng22}.

Due to the importance of rain removal in outdoor vision systems, it has attractted many researchers's attention to solve the problem of restoring rainy images. Single image deraining is more challenging than video-based deraining because there is no inter-frame information.For rain removal from a single image, existing methods fall into two categories: the model-driven method and data-driven method.

{\it Model-driven methods}: Some priors have been employed to remove rain from single images. They assume that rain-streaks $R$ are sparse and in similar directions. They decompose the raining image $O$ into two layers, the rain-free background scene $B$ and the rain-streaks layer $R$.
Kang et al. \cite{1kang2012} proposed an interesting single image deraining algorithm by using the dictionary learning. The rainy image is decomposed into low and high-frequency parts and high-frequency parts of a rainy image are further decomposed into rain and non-rain components. Zhu et al. \cite{zhu} first detected rain-dominant regions and then the detected regions were utilized as a guidance image to help separate rain-streaks $R$ from the background layer $B$.

{\it Data-driven methods}: Data-driven based single image deraining algorithms were popular. Fu et al. \cite{1fu2017} learned the mapping relationship between rainy and clean detail layers, and the predicted detail layer was added into a low-pass filtered base layer for removing rain-streaks. Yang et al. \cite{1yang2017} proposed a deep recurrent dilated network to jointly detect and remove rain-streaks. Zhang et al. \cite{Zhang1} take the rain density into account and present a multi-task CNN for
joint rain density estimation and deraining. Later,  they further improve their work by propose a conditional generative adversarial network for rain-streaks removal in \cite{Zhang2}. Li et al. \cite{1li2018} introduced a novel multi-stage deep learning based single image deraining algorithm. Different $\alpha$ values are assigned to various rain-streak layers according to their properties at each stage, and then a recurrent neural network was incorporated to preserve the useful information in previous stages and benefit the residual prediction in later stage. Data-driven relies on massive amounts of data and is superior to model-driven algorithms in terms of performance. However, even data-driven methods struggle to capture regularity due to the random nature of rainfall.
 
In this paper, a novel rain steaks aware deep convolutional neural network (RSADCNN) for single-image rain removal is designed. The proposed algorithm is based on an observation that prior knowledge can assist data-driven better targeted learning. Since both rain and noise are random, they exist in high frequency components. The high-frequency part can be obtained by an edge-preserving filter such as \cite{1LiIEEETIP2015, 1chen2019}, and then the high-frequency information containing raindrops and noise can effectively guide this data-driven approach to pay more attention to rain or noise in order to recover them efficiently. The RSASCNN incorporates an rain-streaks aware guidance branch (RSAGB) into a Recursive residual group (RRG). The input of RSAGB is high frequency components extracted by an improved weighted guided image filter (iWGIF), and its main components are noise and rain-streaks. The objective of the RSAGB is to help the proposed RSADCNN pay more attention to the rain or noise in order to restore them efficiently. Experimental results validate the prior is indeed very helpful for the single image deraining. The proposed algorithm is useful for autonomous navigation in raining conditions.
 
The rest of this paper is organized as below. The proposed deraining  algorithm is presented in Section \ref{algorithm}. Experimental results are provided in Section \ref{Experiments} to verify the proposed algorithm. Finally, conclusion remarks are drawn in Section \ref{conclusion}

\begin{figure*}[htb]
	\centering
	\includegraphics[width=0.9\textwidth]{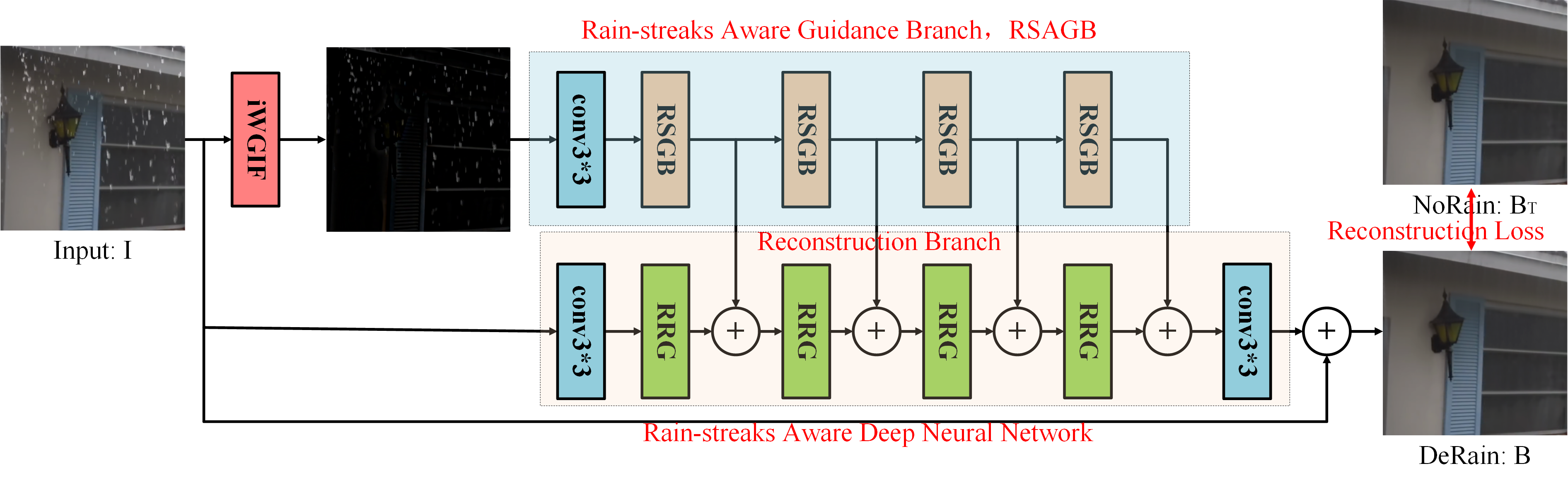}
	\caption{The proposed framework for single-image rain removal. $I$ is a rainy image, $B$ is a restored image, and $B_T$ is the ground truth image of $B$.}
	\label{Fig1}
\end{figure*}

\section{The Proposed Deraining Algorithm}
\label{algorithm}

\subsection{Framework of The Proposed Algorithm}
\label{fundation}

The widely used rainy image model is expressed as \cite{1kang2012}:
{\small \begin{equation}
	\label{eq1}
	I(p)=B(p)+S(p),
	\end{equation}}
where $I$ is a rainy image with rain-streaks; $B$ is the background layer; $S$ is the rain-streak layer; and $p$ is a pixel.  

Single image deraining is to restore the image $B$ from the rainy image $I$. However, when an object’s structure and orientation is similar with that of rain-streaks, it is hard to simultaneously remove rain-streaks and preserve structure, even though DCNN has a strong learning ability. 

A new single image rain-streaks removal algorithm is proposed in this paper by combining an edge-preserving filter and a DCNN. Fig. \ref{Fig1} summarizes the pipeline of our network for the single image derainng algorithm. An improved weighted guided image filter (iWGIF) is first adopted to obtain high frequency components which mainly contain noise, rain-streaks, etc. The high frequency components will be applied to design an RSGB which will guide the RRG to pay more attention to more important regions. The rainy images is then restored by an RSADCNN which is designed by incorporating the RSGB into the RRG.

\begin{figure}[htb]
	\centering
	\includegraphics[width=0.45\textwidth]{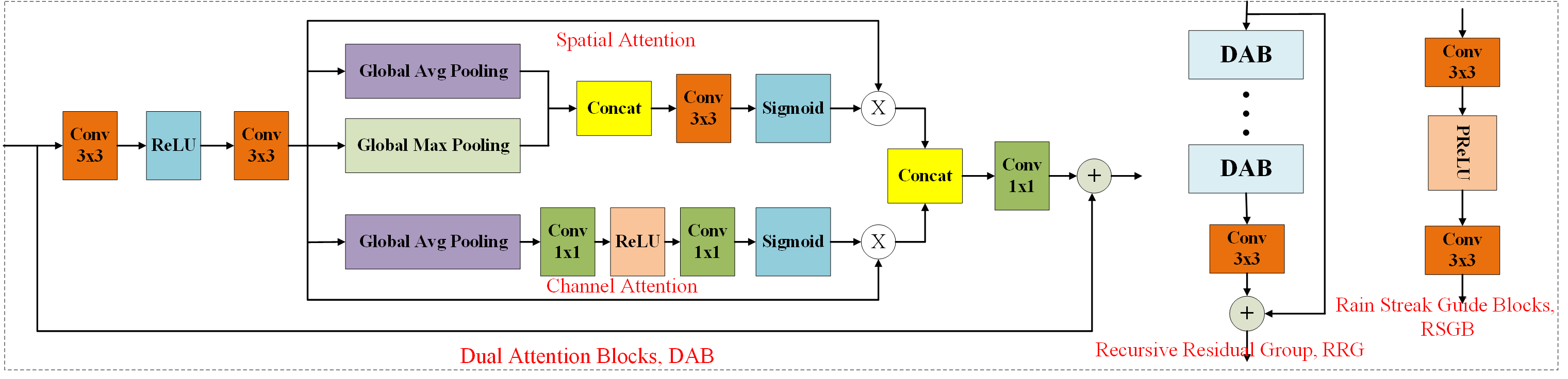}
	\caption{ Recursive residual group (RRG) and Rain Streak Guide Blocks (RSGB),each RRG contains multiple dual attention blocks (DAB). Each DAB contains spatial and channel attention modules.}
	\label{Fig2}
\end{figure}

\subsection{Improved Weighted Guided Image Filtering}

In this subsection, an iWGIF is provided by using the edge-aware weighting in \cite{1LiIEEETIP2015} to improve the WAGIF in \cite{1chen2019}. The objective is to reduce the sensitivity of the WAGIF with respect to the regularization parameter.

Let $I$ be an image to be processing and $G$ be a guidance image.  Let $\Omega_{\zeta}(p)$ be a square window centered at the pixel
$p$ of a radius $\zeta$. Same as  the GIF, WGIF, GGIF and WAGIF,  $I(p)$ is assumed to be a linear transform of the guidance image $G(p)$ in the window $\Omega_{\zeta}(p')$:
\begin{equation}
\label{linearmodel} I(p)=a_{p'}G(p)+b_{p'}, \forall p\in
\Omega_{\zeta}(p'),
\end{equation}
where $a_{p'}$ and $b_{p'}$ are two constants in the
window $\Omega_{\zeta}(p')$.

It should be pointed out that the guidance image $G$ and the image to be processed $I$ can be the same. The optimal values of $a_{p'}$ and $b_{p'}$ are derived by minimizing
a cost function $E(a_{p'},b_{p'})$ which is defined as
\begin{eqnarray}
\label{Eap''bp''old}
\sum_{p\in
	\Omega_{\zeta}(p')}[\Gamma^{G}_{p'}(a_{p'}G(p)+b_{p'}-I(p))^2+\lambda a_{p'}^2],
\end{eqnarray}
where $\lambda$ is a regularization parameter which can be used to penalize a large $a_{p'}$.

Similarly to the WGIF, the edge-aware weighting $\Gamma^{G}_{p'}$ is defined  by using local variances of $3\times 3$ windows of all pixels as follows:
\begin{eqnarray}
\label{w(p)}
\Gamma^{G}_{p'}=\frac{1}{M}\sum_{p=1}^{N}\frac{\sigma^2_{G,1}(p')+\varepsilon}{\sigma^2_{G,1}(p)+\varepsilon},
\end{eqnarray}
where $M$ is the total number of pixels in the image $I$, and  $\varepsilon$ is a small constant. Certainly, there are many different methods to compute the edge-ware weighting \cite{1LiIEEETIP2015}. Different edge-aware weighting could be selected for different applications.

By solving the optimization problem (\ref{Eap''bp''old}), the optimal values of $a_{p'}$ and $b_{p'}$ are computed as
\begin{align}
a_{p'}&=\frac{\Gamma^{G}_{p'}\mbox{cov}_{I,G,\zeta}(p')}{\Gamma^{G}_{p'}\sigma^2_{G, \zeta}(p')+\lambda},\\
b_{p'}&=\mu_{I,\zeta}(p')-a_{p'}\mu_{G,\zeta}(p'),
\end{align}
where the operation $\odot$ is the element-by-element product of two matrices. $\mbox{cov}_{I,G,\zeta}(p')$ is
\begin{align}
\mbox{cov}_{I,G,\zeta}(p')=\mu_{G\odot I, \zeta}(p')-\mu_{G,\zeta}(p')\mu_{I,\zeta}(p'),
\end{align}
and $\mu_{G, \zeta}(p')$, $\mu_{I, \zeta}(p')$, and  $\mu_{G\odot I, \zeta}(p')$ are the mean values of $G$, $I$ and $G\odot I$ in the window $\Omega_{\zeta}(p')$,  respectively.

Instead of using the averaging method in \cite{1he2013,1LiIEEETIP2015}, an optimal based method is adopted to compute the final value of $I(p)$ as
\begin{equation}
\min_{I(p)}\{\sum_{p'\in \Omega_{\zeta}(p)}W_{p'}(I(p)-a_{p'}G(p)-b_{p'})^2\},
\end{equation}
where $W_{p'}$  is a weighting factor and it is given as \cite{1chen2019}
\begin{align}
W_{p'}&=\exp^{-\frac{1}{|\Omega_{\zeta}(p')|}{\displaystyle \sum_{p\in \Omega_{\zeta}(p')}}\frac{(a_{p'}G(p)+b_{p'}-I(p))^2}{\eta}}+0.001,
\end{align}
$\eta$ is a small positive constant, and $|\Omega_{\zeta}(p')|$ is the cardinality of the set $\Omega_{\zeta}(p')$.

The optimal solution is computed as
\begin{equation}
I^*(p)=\bar{a}_p^WG(p)+\bar{b}_p^W,
\end{equation}
where $\bar{a}_p$ and $\bar{b}_p$ are computed as
\begin{align}
\bar{a}_p^W& =\frac{1}{W^{sum}_p}\sum_{p'\in \Omega_{\zeta}(p)}W_{p'}a_{p'},\\
\bar{b}_p^W& =\frac{1}{W^{sum}_p}\sum_{p'\in \Omega_{\zeta}(p)}W_{p'}b_{p'},
\end{align}
and $W_p^{sum}$ is
\begin{align}
W_p^{sum}&=\sum_{p'\in \Omega_{\zeta}(p)}W_{p'}.
\end{align}

It can be easily derived that
\begin{align}
\nonumber
&\frac{1}{|\Omega_{\zeta}(p')|}\sum_{p\in \Omega_{\zeta}(p')}(a_{p'}G(p)+b_{p'}-I(p))^2\\\nonumber
=&a_{p'}^2\sigma_{G,\zeta}^2(p')-2a_{p'}\mbox{cov}_{I,G,\zeta}(p')+\sigma_{I,\zeta}^2(p')\\
=&\sigma_{I,\zeta}^2(p')-a_{p'}^2(\sigma_{G,\zeta}^2(p')+\frac{2\lambda}{\Gamma_{p'}^G}).
\end{align}

Therefore, same as the GIF in \cite{1he2013}, WGIF in \cite{1LiIEEETIP2015} and WAGIF in \cite{1chen2019}, the complexity of the proposed iWGIF is $O(M)$ for an image with $M$ pixels.

They are many applications of the proposed iWGIF. For example, it can be applied to study the problems in \cite{2kouf2015}. When the proposed iWGIF is adopted to decompose the rainy image $I$ into two layers as in the equation (\ref{eq1}), the guidance image $G$ is the same as the rainy image $I$

\subsection{Rain-streaks aware deep convolutional neural network }
Both noise and rain-streaks are categorized as high frequency components, respectively. Instead of performing end-to-end learning directly on the input rainy images and clear images, we can use it to guide the network learning, then the learning will be more efficient. Fig. \ref{Fig1} summarizes the pipeline of our network for the Single Image Derainng algorithm. The CNN used in the proposed framework is on top of the DeNoiseNet in \cite{CycleISP}, which achieves outstanding performance in denoising. The recursive residual group (RRG) in DeNoiseNet is shown in Fig. \ref{Fig2}. The function of the RRG is to globally suppress the less useful features and only allow the propagation of more informative ones. Firstly, the RRG contains Dual Attention Blocks (DAB) and each DAB performs both spatial and channel attention operations. Since the shape and direction of raindrops are randomly changed during the falling, the attention mechanisms treat different features and pixels region unequally, which can provide additional flexibility in dealing with different types of information, and can expand the representational ability of CNNs. Secondly, the proposed RSADCNN contains several Rain Streak Guide Blocks (RSGB), the input of RSGB the high frequency information, which focus on rain-streaks and noise, and directly guide the network to pay more attention to rain-streaks from another angle.

Loss functions play an important role in training the CNN from $N$ pairs of images $\{(I,B_T)\}$. Here, $I$ is a rainy image and $B_T$ is a  rain-free image. It is well known that $L_2$ penalizes larger errors and tolerant to small errors, while $ L_1$ does not over-penalize large errors. They have different convergence properties. The new reconstruction loss function $L_r$ is a hybrid $L_1/L_2$ norm which considers the advantages of these two loss functions. It is defined as
\begin{equation}
\label{lrlr}
L_r = \sum_p\psi(B_T - B),
\end{equation}
where the function $\psi(z)$ is defined as \cite{1wangwei2019}
\begin{equation}
\psi(z)=\left\{\begin{array}{ll}
|z|;&\mbox{if~}|z|>c\\
\frac{z^2+c^2}{2c}; &\mbox{otherwise}\\
\end{array}
\right.,
\end{equation}
and $c$ is a positive constant and its value is selected as 2/255 in this paper. 

The above function is derived from the well known Huber loss function. It is easily shown that the function  $\psi(z)$ is differentiable. Let  $\psi'(z)$ be the derivative of the function $\psi(z)$, and it is clearly a continuous function given as:
\begin{equation}
\psi'(z)=\left\{\begin{array}{ll}
1; &\mbox{if~}z\geq c\\
-1; &\mbox{if~}z\leq -c\\
\frac{z}{c}; &\mbox{otherwise}\\
\end{array}
\right..
\end{equation}

\section{Experiments}
\label{Experiments}
In this section, we first study the effect of RSGB, then evaluate the effectiveness of the proposed method on the existing deraining datasets,  compare the proposed deraining method  with several state-of-the-art single-image deraining methods.  Readers are invited to view to electronic version of figures and zoom in them so as to better appreciate differences among all images.

\subsection{Implementation Details}
To validate and evaluate our method, we conduct the comparison and analysis experiments on synthetic rain image Rain100L \cite{DJRain}, Rain100H \cite{DJRain} and Rain1400 \cite{DDRain}. PSNR and SSIM are choosed as the evaluation metrics.

The number of RRGs and RSGB are set as $4$, and each RRG contains $8$ DABs. We randomly crop a $128*128$ patch from each input during training. The proposed network is trained using the proposed loss functions and Adam optimizer with ${{\beta}_1}=0.9$ and ${{\beta}_2}=0.99$. We set the batch size to $16$. The learning rate is initially set to $10^{-4}$ and then decreased using a cosine annealing schedule. All the experiments are implemented using PyTorch on NVIDIA GP100 GPUs.

\subsection{Analysis of RSGB}

\begin{figure}[htb]
	\centering
	\subfigure{
		\begin{minipage}[b]{0.48\linewidth}
			\includegraphics[width=1\linewidth]{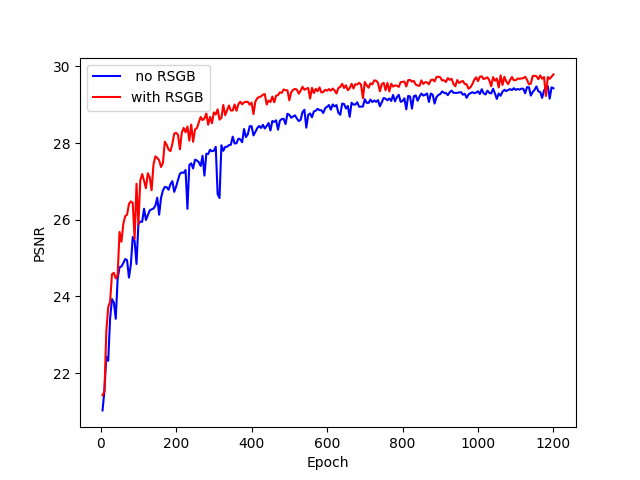}	
			\centerline{(a)}	
	\end{minipage}}
	\subfigure{
		\begin{minipage}[b]{0.48\linewidth}
			\includegraphics[width=1\linewidth]{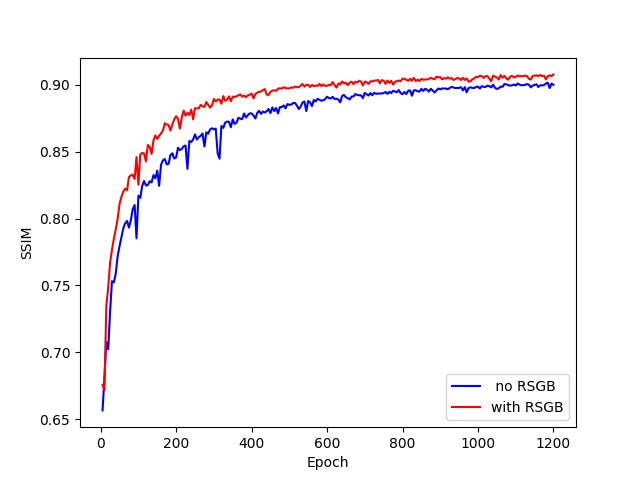}
			\centerline{(b)}
	\end{minipage}}
	\caption{(a) Comparisons of PSNR between the modes with and without the RSGB for Rain100H images. (b) Comparisons of SSIM between the modes with and without the EAGB for Rain100H images. The mode with the RSGB is more stable than the mode without RSGB, and can also achieve higher PSNR and SSIM. }
	\label{iWGIF}
\end{figure}

 The features from the RSGB are embedded into the RRG to guide the RRG.  In order to validate the effectiveness of the RSGB,  the two modes with and without the RSGB are compared. As illustrated in Table \ref{ta2b}, both the PSNR and the SSIM are improved by using the RSGB. In addition,  the mode with the RSGB is more stable than the mode without EAGB, and can achieve higher PSNR and SSIM values in different epoches as shown in Fig. \ref{iWGIF}. Clearly, the RSGB indeed guides the RRG to exploit rain Streaks information for restoring the synthetic images.

\begin{table}[htb]
	\centering
	\begin{threeparttable}
		\caption{SSIM and PSNR of synthetic images generated by different methods for Rain100L and Rain100H.}
		\begin{tabular}{ccccccc}
			\toprule
			\multirow{2}{*}{}&
			\multicolumn{2}{c}{ SSIM }&\multicolumn{2}{c}{ PSNR }\cr
			\cmidrule(lr){2-3} \cmidrule(lr){4-5}
			& Rain100L  &  Rain100H &  Rain100L & Rain100H\cr
			\midrule
			(no RSGB)            &0.9852   &0.9033   &38.79    &29.57     \cr
			proposed             &\textbf{0.9863}  &\textbf{0.9082}   &\textbf{39.11}   &\textbf{29.78}    \cr
			\bottomrule
		\end{tabular}
		\label{ta2b}
	\end{threeparttable}
\end{table}

\begin{figure*}[htb]
	\centering
	\includegraphics[width=0.90\textwidth]{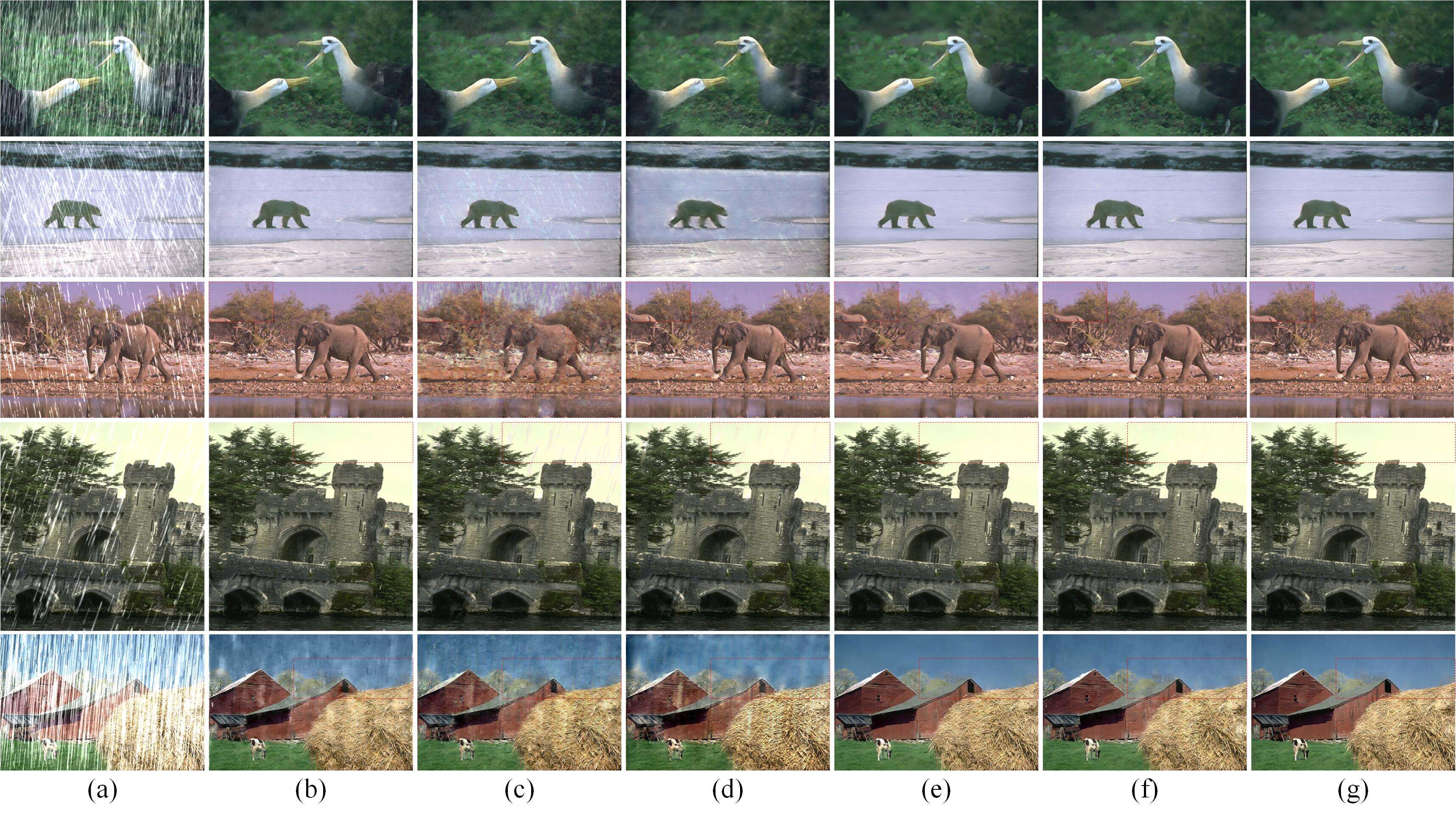}
	\caption{Comparison of different deraining algorithms on synthetic rainy images. From left to right, the rainy images and the restored images by  \cite{DJRain},  \cite{DDRain},  \cite{LPNet},  \cite{RCDNet}, \cite{VRGNet} and the proposed method, respectively. }
	\label{Fig4}
\end{figure*}

\begin{table}[htb]
	\vspace{-3mm}
	\caption{SSIM and PSNR Of Different Algorithms for Rain100L}
	\centering
	{\scriptsize\begin{tabular}{ c|ccccccccc }
			\hline
			& \cite{DJRain}&\cite{DDRain}&\cite{LPNet}& \cite{RCDNet}& \cite{VRGNet} & Ours \\
			\hline
			SSIM &0.965 &0.948     &0.952  &0.984  &0.98     &\textbf{0.986}   \\
			PSNR &34.32 &32.94    &32.15   &38.58  &38.21    &\textbf{39.11}  \\
			\hline	
	\end{tabular}}
	\label{tabfusion1}
\end{table}

\begin{table}[htb]
	\vspace{-3mm}
	\caption{SSIM and PSNR Of Different Algorithms for Rain100H}
	\centering
	{\scriptsize\begin{tabular}{ c|ccccccccc }
			\hline
			& \cite{DJRain}&\cite{DDRain} &\cite{LPNet}& \cite{RCDNet}& \cite{VRGNet} & Ours \\
			\hline
			SSIM &0.786 &0.751   &0.760  &0.886  &0.855   &\textbf{0.908}   \\
			PSNR &24.37 &24.56   &21.79  &28.81  &27.62    &\textbf{29.78}  \\
			\hline	
	\end{tabular}}
	\label{tabfusion2}
\end{table}

\begin{table}[htb]
	\vspace{-3mm}
	\caption{SSIM and PSNR Of Different Algorithms for Rain1400}
	\centering
	{\scriptsize\begin{tabular}{ c|ccccccccc }
			\hline
			& \cite{DJRain}&\cite{DDRain} &\cite{LPNet}& \cite{RCDNet}& \cite{VRGNet} & Ours \\
			\hline
			SSIM &0.876   &0.864   &0.877  &0.909  &0.908   &\textbf{0.915}   \\
			PSNR &28.37   &28.19   &28.21  &30.59  &30.6    &\textbf{31.23}  \\
			\hline	
	\end{tabular}}
	\label{tabfusion3}
\end{table}

\subsection{Comparison with Existing Algorithms}

The proposed algorithm is compared with five state-of-the-art deraining algorithms in \cite{DJRain},  \cite{DDRain},  \cite{LPNet},  \cite{RCDNet}, \cite{VRGNet} on synthetic rainy dataset. These algorithms were published at top conferences in computer vision. As shown in Fig. \ref{Fig4},  the algorithms in \cite{VRGNet, RCDNet} and the proposed algorithm can be adopted to reduce rain-streaks better than the algorithms in \cite{DJRain, DDRain, LPNet}. There are still visible artifact in the restored images by the algorithm in \cite{VRGNet, RCDNet}, as shown in the red box. It is worth noting that the algorithm in \cite{DDRain} also directly uses high-frequency components to learn raindrops to avoid background interference. Due to the randomness of raindrop, high-frequency information cannot completely cover raindrop information, and the efficiency  is limited.
 
Besides the subjective evaluation, the objective quality metrics including the SSIM and PSNR are adopted to further compare the proposed algorithm with those in \cite{DJRain},  \cite{DDRain},  \cite{LPNet},  \cite{RCDNet}, \cite{VRGNet}.  The average SSIM and PSNR are shown in Tables \ref{tabfusion1}-\ref{tabfusion3}. The results further prove the superiority of the proposed algorithm from the objective quality point of view.

\section{Conclusion Remarks and Discussions}
\label{conclusion}
A new data-driven deraining algorithm is proposed by utilizing prior knowledge which is derived from a fast edge-preserving filter. Under the guidance of  the prior knowledge, the network  pays more attention to more important regions. As such, the learning efficiency will be higher. Experimental results show that the proposed algorithm could outperform several existing single image de-raining algorithms.

For other low-level visual processing tasks such as dehazing and illumination enhancement, the prior knowledge can also be incorporated into data-driven approaches. Besides the prior knowledge, it is also important to combine model-based and data-driven methods for the low-level visual processing. The proposed algorithm can also be applied to study autonomous navigation in raining conditions. All these problems will be studied in our future research.


\begin{thebibliography}{1}
	
	
\bibitem{1Nara2000} S. G. Narasimhan and S. K. Nayar,
\newblock ``Chromatic framework for vision in bad weather,''
\newblock In {\it Proc. IEEE Conf. Computer Vision and Pattern Recognition (CVPR)}, vol. 1, Jun. 2000, pp. 598-605.

\bibitem{1greg2004}K. Garg and S. K. Nayar,
\newblock ``Detection and removal of rain from videos."
\newblock in {\it IEEE CVPR}, vol. 1, pp. 528-535, Jun. 2004.

\bibitem{2li2021} Z. G. Li, H. Y. Shu, and C. B. Zheng, 
\newblock ``Multi-scale single image dehazing using Laplacian and Gaussian pyramids,"
\newblock IEEE Trans.  on Image Processing, vol. 30, pp. 9270-9279, Dec. 2021.


\bibitem{zheng22} C. B. Zheng, Z. G. Li, Y. W. Li and S. Q. Wu, 
\newblock ``Non-Local Single Image DE-Raining Without Decomposition,"
\newblock  in {\it IEEE International Conference on Acoustics, Speech and Signal Processing}, pp. 1425-1429, 2021.


\bibitem{1kang2012}
L.W. Kang,  C. W. Lin, and  Y. H. Fu,
\newblock ``Automatic single-image-based rain streaks removal via image decomposition,"
\newblock  IEEE Trans. on Image Processing , vol. 21, no. 4, pp. 1742-1755, Apr. 2012.

\bibitem{zhu} L. Zhu, C. W. Fu, D. Lischinski, and P. A. Heng,
\newblock ``Joint bi-layer optimization for single-image rain streak removal,"
\newblock  in {\it IEEE International Conference on Computer Vision}, pp. 2545–2553, 2017.

\bibitem{Zhang1} H. Zhang and V. M. Patel,
\newblock ``Density-aware single image deraining using a multi-stream dense network,"
\newblock  in {\it IEEE Conference on CVPR}, pp. 695–704, 2018.


\bibitem{Zhang2}
H. Zhang, V. Sindagi, and V. M. Patel,
\newblock ``Image deraining using a conditional generative adversarial network,"
\newblock  IEEE Trans. on Circuits and Systems for Video Technology , vol. 30, no. 11, pp. 3943 - 3956, Nov. 2020.

\bibitem{1fu2017} X. Fu, J. Huang,  X. Ding,  Y. Liao, and J. Paisley,
\newblock ``Clearing the skies: a deep network architecture for single-image rain removal,"
\newblock  IEEE Trans. on Image Processing, vol. 26, no. 6, pp. 2944-2956, Jun. 2017.

\bibitem{1yang2017} W. Yang,  R. T. Tan, J. Feng, J. Liu, Z. Guo,  and S. Yan,
\newblock ``Deep joint rain detection and removal from a single image,"
\newblock in {\it IEEE CVPR}, pp. 1357-1366, 2017.

\bibitem{1li2018} X. Li, J. Wu, Z. Lin, H. Liu, and H. Zha,
\newblock ``Recurrent squeeze-andexcitation context aggregation net for single image deraining,"
\newblock in {\it ECCV}, pp. 254-269, 2018.


\bibitem{DJRain} W. Yang, R. T. Tan, J. Feng, J. Liu, Z. Guo, and S. Yan.
\newblock ``Deep joint rain detection and removal from a single image,"
\newblock in {\it IEEE CVPR}, pp. 1685-1694, 2017.

\bibitem{DDRain} X.Y. Fu, J. B. Huang, D. L. Zeng, Y. Huang, X. H. Ding, and J. W. Paisley.
\newblock ``Removing rain from single image via a deep detail network,"
\newblock in {\it IEEE CVPR}, pp. 1715-1723, 2017.

\bibitem{LPNet} X.Y. Fu, B. R. Liang, Y. Huang, X. H. Ding, and J. Paisley.
\newblock ``Lightweight Pyramid Networks for Image Deraining,"
\newblock  IEEE Transactions on Neural Networks and Learning Systems, vol. 31, no. 6, pp. 1794-1807, June. 2020.


\bibitem{RCDNet}
H. Wang, Q. Xie. Wang, Q. Zhao, and D. Y. Meng,
\newblock ``A Model-Driven Deep Neural Network for Single Image Rain Removal,"
\newblock in {\it IEEE CVPR}, 2020.

\bibitem{VRGNet} H. Wang, Z. S. Yue, Q. Xie, Q. Zhao, Y,F, Zheng, and  D. Y. Meng
\newblock ``From Rain Generation to Rain Removal,"
\newblock in {\it IEEE CVPR}, pp. 14791-14801, 2021.


	
\bibitem{1he2013} K. He, J. Sun, and X. Tang,
\newblock ``Guided image filtering,"
\newblock  IEEE Trans. on Pattern Analyis and Machine Intelligence, vol. 35, no. 6, pp. 1397-1409, Jun. 2013.

 \bibitem{2kouf2015} F. Kou, W. H. Chen, Z. G. Li, and C. Y. Wen,
 \newblock ``Content adaptive image detail enhancement,"
 \newblock IEEE Signal Processing Letters, vol. 22,  no. 2,  pp. 211-215,  Feb. 2015.


\bibitem{CycleISP} S. Zamir, A. Arora, S. Khan, M. Hayat, F. S. Khan, M. Yang, and L. Shao
\newblock ``CycleISP: real image restoration via improved data synthesis,"
\newblock in {\it IEEE CVPR}, 2020.

\bibitem{1LiIEEETIP2015} Z. G. Li, J. H. Zheng, Z. J. Zhu, W. Yao, and S. Q. Wu,
\newblock ``Weighted guided image filtering,"
\newblock IEEE Trans. on Image Processing. vol. 24, no.1, pp. 120-129, Jan. 2015.

\bibitem{1wangwei2019}	W. Wang, Z. G. Li, S. Q. Wu, and L. C. Zeng,
\newblock ``Haze image decolorization with color contrast restoration,"
\newblock IEEE Trans.  on Image Processing, vol. 28, no. 12, pp. 1776 - 1787,  Dec. 2019.

\bibitem{1chen2019} B. Chen and S. Q. Wu,
\newblock ``Weighted aggregation for guided image filtering,"
\newblock Signal, Image and Video Processing, pp. 1-8,  Oct. 2019.

\end{thebibliography}
\end{document}